\documentclass[a4paper,fleqn]{cas-dc}

\usepackage[numbers]{natbib}
\usepackage{xcolor}
\usepackage{amsmath,amsfonts}
\usepackage{algorithmic}
\usepackage{algorithm}
\usepackage{array}
\usepackage[caption=false]{subfig}
\usepackage{textcomp}
\usepackage{stfloats}
\usepackage{url}
\usepackage{verbatim}
\usepackage{graphicx}
\usepackage{multirow}
\usepackage{hyperref}
\usepackage{threeparttable}

\def\tsc#1{\csdef{#1}{\textsc{\lowercase{#1}}\xspace}}
\tsc{WGM}
\tsc{QE}
\tsc{EP}
\tsc{PMS}
\tsc{BEC}
\tsc{DE}

\begin{document}
\let\WriteBookmarks\relax
\def\floatpagepagefraction{1}
\def\textpagefraction{.001}
\shorttitle{SPRUNN}
\shortauthors{B. Rokh et~al.}

\title [mode = title]{\textbf{SPRUNN}: \textbf{S}tructural \textbf{P}roximity for P\textbf{ru}ning in Deep \textbf{N}eural \textbf{N}etworks}                      



\author[1]{Babak Rokh}[orcid=0000-0003-2481-5362]
\ead{babak.rokh@znu.ac.ir}

\author[1]{Ali Azarpeyvand}[orcid=0000-0002-4166-7528]
\cormark[1]
\ead{azarpeyvand@znu.ac.ir}

\author[2]{Alireza Khanteymoori}[orcid=0000-0001-6811-9196]
\ead{alireza.khanteymoori@psychologie.uni-freiburg.de}

\author[3]{Tara Ghasempouri}[orcid=0000-0001-8021-9368]
\ead{tara.ghasempouri@taltech.ee}

\affiliation[1]{organization={Department of Computer Engineering, University of Zanjan},
                city={Zanjan},
                country={Iran}}

\affiliation[2]{organization={Department of Psychology, University of Freiburg},
                city={Freiburg},
                country={Germany}}

\affiliation[3]{organization={Department of Computer Systems, Tallinn University of Technology},
                city={Tallinn},
                country={Estonia}}

\cortext[cor1]{Corresponding author}


\begin{abstract}
Deep Neural Networks (DNNs) have demonstrated significant accuracy across various machine learning tasks. However, these networks suffer from a huge number of parameters and computations, which leads to time-consuming processes. Consequently, parameter pruning approaches have been proposed to compress DNNs and efficiently implement them. The main challenge in pruning is selecting the appropriate parameters for pruning without significantly damaging the accuracy. In this paper, we propose SPRUNN, an algorithmic pruning approach based on the \textbf{s}tructural \textbf{p}roximity criterion for filter p\textbf{ru}ning in convolutional layers of deep \textbf{n}eural \textbf{n}etworks. Structural Proximity (SP) calculates filter similarity by considering all filters and comparing each one with the others, effectively identifying less important filters for pruning in the convolutional layers. Furthermore, we integrate this approach with fine-grained pruning in Fully Connected (FC) layers to compress the DNNs that include both convolutional and FC layers. To enhance the accuracy of the pruned model, we propose a learning rate refinement strategy during the retraining phase. Experimental results demonstrate that SPRUNN achieves significant compression with high accuracy across diverse architectures and datasets.
\end{abstract}

\begin{keywords}
Deep learning \sep Model compression \sep Network pruning \sep Filter similarity
\end{keywords}

\maketitle

\section{Introduction}
Recent advances in machine learning tasks using Deep Neural Networks (DNNs) are considerable \cite{wani2025advances}. However, the huge number of parameters in DNNs creates a bottleneck in processing speed during inference, which leads to time-consuming and high-cost power in transferring data between memories as well as computations. Therefore, DNN compression has been a trending topic in the last decade. Among various compression techniques, pruning has gained considerable attention in recent years due to its effectiveness in accelerating the inference of DNNs \cite{han2015learning, he2018soft, he2019filter, lin2021filter, he2022filter, liu2023eacp, shi2024sparse, khan2025pruning, ZHANG2026134207, guenter2026concurrent}. 

The main challenge in pruning is identifying the weights that can be removed while preserving model accuracy. Consider a network with $n$ weights. Since each weight can either be retained or pruned, the number of possible pruning configurations is $2^n$, implying an exponential search complexity for finding the optimal pruning configuration. To address this challenge, most pruning methods rely on criterion-based approaches, where a predefined criterion is used to estimate the importance of weights and identify candidates for removal. Some criterion-based methods use the magnitude of the weights to identify candidate weights for pruning \cite{he2018soft, plochaet2023hardware, 10103462, shi2024sparse, tyche2024improving}, while some of them utilize similarity criteria for pruning \cite{he2019filter, liu2023eacp, zu2023consecutive, wang2023progressive, li2025sfp}. In convolutional layers, all filters are convolved with the same input feature maps. Consequently, when some filters are highly similar to others, their removal can often be compensated for by the remaining filters, since the features extracted by the pruned filters can still be represented by the retained ones. This characteristic makes similarity-based pruning particularly effective for convolutional layers. Previous studies have proposed similarity measures such as Geometric Median (GM) \cite{he2019filter} and Cosine similarity \cite{ZHANG2023116, zu2023consecutive, chang2023iterative} to identify redundant filters within a layer. However, these criteria have limitations in accurately identifying less important filters. GM primarily focuses on the global similarity among filters and may neglect local structural relationships, whereas Cosine criterion evaluates only the directional alignment of filters and does not fully capture their structural similarity.

Another important challenge in pruning is maintaining the accuracy of the pruned model. To mitigate the accuracy degradation caused by pruning, several studies have proposed retraining the network after pruning \cite{han2015learning, lin2021filter, lin2021network, lee2022dnn, kamma2023pruning}. However, effectively adjusting the model parameters during retraining remains a significant challenge. Most prior works typically perform basic fine-tuning following pruning to recover potential accuracy loss \cite{renda2020comparing}, where the final parameter values from the training phase are used as the initial values in retraining the pruned model. Reference \cite{frankle2020linear} instead propose weight rewinding for retraining. In this approach, the unpruned weights are initialized using their values from earlier epochs of the original training. The results reported in \cite{renda2020comparing} demonstrate that initializing retraining with the final trained weights yields better performance compared to rewinding the weights to values from $t$ epochs earlier in the training process. Building upon weight rewinding, reference \cite{renda2020comparing} introduces learning rate rewinding, where, instead of resetting the learning rate to its initial value during retraining, it is reset to a value from $t$ epochs prior to the final epoch of the original training.

In this paper, we propose SPRUNN, an algorithmic pruning approach that introduces the Structural Proximity (SP) criterion as a similarity measure to identify redundant filters in convolutional layers. Unlike existing similarity measures, SP evaluates similarity among filters by comparing the elements of each filter with the corresponding elements of all other filters within the same layer, enabling more accurate identification of redundant filters for pruning. Additionally, we also employ fine-grained magnitude-based pruning in the Fully Connected (FC) layers, combining it with the approach used in convolutional layers to achieve significant compression while maintaining accuracy in DNNs with both convolutional and FC layers. We first train the network, and after convergence, candidate filters and weights are identified for pruning. After one-shot pruning, the pruned model undergoes retraining to optimize the remaining weights. During retraining, we suggest learning rate refinement to effectively manage exploration and improve the accuracy of the pruned model. The main contributions of this paper are summarized as follows:
\begin{itemize}
    \item \textbf{Structural Proximity Criterion:} We propose a novel similarity criterion, termed SP, for identifying appropriate candidates for filter pruning. SP computes the similarity between filters by comparing the elements of each filter to those of all other filters in the same layer, ensuring that the remaining filters can adequately compensate for those removed. The introduced criterion demonstrates acceptable time complexity, making it computationally efficient and scalable to DNNs with a large number of filters, thus enabling practical applicability.
    \item \textbf{Multi-Granularity Pruning Framework:} The proposed SPRUNN framework integrates similarity-based coarse-grained structured pruning for convolutional layers with magnitude-based fine-grained pruning for FC layers. This multi-granularity approach leverages the effectiveness of similarity-based filter pruning in convolutional layers and the precision of weight-level pruning in FC layers, achieving improved compression with minimal accuracy degradation.
    \item \textbf{Learning Rate Refinement:} We suggest a learning rate refinement strategy for retraining the pruned model. By adjusting the learning rate to a suitably small value, we effectively control the exploration behavior of the optimization process. This refinement of learning rate leads to improved accuracy in the pruned model. 
    \item \textbf{Extensive Empirical Evaluation:} The results show that SPRUNN effectively prunes DNNs while maintaining high accuracy. The proposed method achieves competitive or superior performance compared with state-of-the-art (SOTA) pruning methods across various architectures and datasets. For example, on ResNet-110, SPRUNN achieves an accuracy of 91.75\% on CIFAR-10 while removing 95.63\% of the parameters. For MobileNetV2, it attains an accuracy of 93.22\% with a 90.72\% reduction in parameters. By incorporating pruning in the FC layers, SPRUNN also yields competitive results on LeNet-5 and VGG-16. Specifically, on VGG-16, it achieves a 96.33\% reduction in parameters while maintaining an accuracy of 93.40\% on the CIFAR-10 dataset. Moreover, the proposed method demonstrates a effective trade-off between accuracy and compression on the challenging fine-grained CUB-200-2011 dataset, achieving 71.63\% accuracy while removing 74.75\% of the model parameters. In addition, we conduct a comprehensive analysis studies to evaluate the individual contribution and effectiveness of each component of the SPRUNN framework.    
\end{itemize}
The rest of the paper is organized as follows. In Section \ref{sec_related_works}, we investigate existing approaches on pruning. Our methodology is thoroughly described in Section \ref{sec_method}. In Section \ref{sec_results}, we present and discuss the experimental results and compare SPRUNN with SOTA methods. Section \ref{sec_discussion} provides a summary discussion of the efficiency of the proposed method, while Section \ref{sec_conclution} presents the conclusions and recommendations for future work.

\section{Related works}\label{sec_related_works}
Previous works in pruning can be discussed from different aspects. In this section, we elaborate on various approaches to pruning.

\subsection{Unstructured and structured pruning}
In unstructured pruning, the candidate weights for pruning are removed without considering the model structure, while in structured pruning, all weights in a specific structure are removed \cite{yu2022hessian}. The unstructured approach more precisely prunes less important weights, typically enabling higher pruning rates with minimal accuracy degradation. However, it leads to irregular sparsity and presents challenges for DNN implementation on hardware, where specialized hardware is required. Because of this, the unstructured approach does not significantly improve computational performance unless in high pruning rates \cite{park2016faster}. References \cite{han2015learning, shi2024sparse} are some samples of unstructured pruning works. In contrast, structured pruning removes weights in a regular pattern, making it suitable for implementation on DNN hardware accelerators without requiring specialized hardware \cite{plochaet2023hardware}. However, this approach may inadvertently remove some important connections within the structured pattern. Consequently, maintaining model accuracy during structured pruning can be more challenging compared to unstructured pruning. Since structured pruning algorithms are designed based on the network architecture, they may not apply to the various types of networks. References \cite{he2018soft, he2019filter, lin2021filter, he2022filter, chen2022fpc, ghimire2023loss, plochaet2023hardware, liu2023eacp, ZHANG2023116, zu2023consecutive, wang2023progressive, gupta2024torque, li2025sfp, lian2025daar, tofigh2026resource} propose filter pruning and References \cite{kim2023trunk, tyche2024improving, pinto2026towards} suggest channel pruning. 

Filter pruning leads to a significant reduction in the number of parameters and floating point operations (FLOPs) in Deep Convolutional Neural Networks (DCNNs). Since the number of filters in a given layer determines the number of input channels in the subsequent layer, pruning filters in one layer not only removes those filters but also eliminates the corresponding input channels, thereby decreasing parameters in the following layer. Some previous methods focusing on filter or channel pruning, such as \cite{he2018soft, he2019filter, he2022filter, zheng2024novel}, do not address pruning in the FC layers, limiting the compression potential of DCNNs that include such layers. In this work, we propose a multi-granularity pruning, in which a coarse-grained structured filter pruning method for convolutional layers is effectively integrated with fine-grained pruning for FC layers, enabling efficient compression of DNNs with FC layers.

\subsection{Criterion-based pruning}
Most pruning methods are criterion-based. In the criterion-based approach, candidate weights for pruning are selected based on their values. The most popular criteria for pruning are computed based on the magnitude and similarity of the weights. In magnitude-based pruning approaches, the belief is that the smallest weights are the best candidates for pruning, while similarity-based approaches concentrate on the similarity measures. The idea behind magnitude-based methods arises from the observation that small weights have less contribution than large weights to accuracy. 

References \cite{he2018soft, plochaet2023hardware, 10103462, shi2024sparse, tyche2024improving} suggest magnitude-based pruning. The Soft Filter Pruning (SFP) method \cite{he2018soft} prunes filters with smaller $l_p$-norm. Reference \cite{plochaet2023hardware} suggests a magnitude-based filter pruning, where the importance of each channel throughout the entire network is determined using normalized $l_2$-norm. Dynamic Processing Unit (DPU) \cite{10103462} proposes a dynamic pruning, where candidate weights for pruning are determined based on magnitude. Sparse Optimization Guided Pruning (SOGP) \cite{shi2024sparse} suggests a magnitude-based pruning approach. Instead of directly applying a criterion to specify candidate weights for pruning, SOGP determines these weights in the loss function by introducing fine-grained sparse regularization.

References \cite{he2019filter, wang2023progressive, liu2023eacp, zu2023consecutive, ZHANG2023116, khan2025pruning, li2025sfp} propose a similarity-based pruning. Filter Pruning via Geometric Median (FPGM) \cite{he2019filter} suggests pruning redundant filters using the GM as a similarity criterion, in which filters closest to the GM are considered redundant and subsequently pruned. Effective
Automatic Channel Pruning (EACP) \cite{liu2023eacp} first clusters filters using k-means++, then preserves the filter with the smallest distance to the cluster center while pruning the remaining filters in each cluster. Progressive
Local Filter Pruning (PLFP) \cite{wang2023progressive} introduces a local filter pruning criterion that computes the similarity of a filter with its $k$ nearest neighbors, rather than considering all filters. Reference \cite{ZHANG2023116} performs filter pruning based on the uniqueness of the feature maps, where feature maps are first transformed into the frequency domain using a Discrete Cosine Transform. Then, feature map uniqueness is computed based on the correlation, and feature maps with low uniqueness, as well as their corresponding filters, are removed. Consecutive Layer Collaborative filter Similarity (CLCS) \cite{zu2023consecutive} suggests filter pruning based on the Cosine similarity criterion, measuring the similarity of filters in two consecutive layers rather than within a single layer.

Considering the weight-sharing mechanism and the fact that each filter convolves over all input feature maps in convolutional layers, we argue that similarity-based criteria are particularly well-suited for filter pruning. However, existing similarity criteria exhibit limitations in effectively identifying truly redundant filters. For instance, methods based on the GM compute similarity globally and prune filters that are closest to a central geometric point. However, the GM criterion may miss subtle critical local structural variations within the filter space. Similarly, Cosine criterion identifies filters with similar directional orientation but may fail to capture differences in magnitude or element-wise distribution. In this paper, to address these limitations, we adopt Crowding Distance from the Nondominated Sorting Genetic Algorithm II (NSGA-II) \cite{996017} and introduce a novel similarity criterion, SP, to identify and prune less important filters in convolutional layers. The SP criterion quantifies similarity based on the relative distribution of filter elements, offering a more comprehensive measure of structural redundancy.
\begin{figure*}
\centering
\includegraphics[scale=0.7]{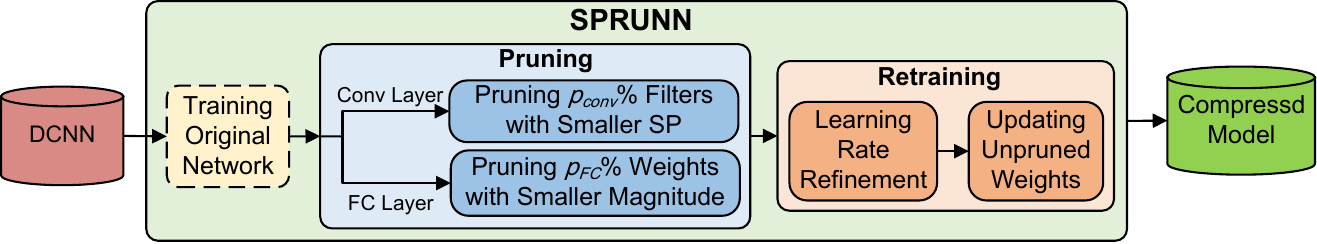}
\caption{Overview of the steps in SPRUNN (The step of training original networks is omitted when a pretrained model is given to SPRUNN).}
\label{steps}
\end{figure*}

\subsection{During-training and post-training pruning}
Some prior works integrate weight pruning into the training process from scratch, so that model optimization and pruning occur simultaneously. Conversely, some works train networks with all the weights initially, and after the unpruned model converges, pruning is performed, followed by retraining to fine-tune the remaining weights. 

References \cite{he2018soft, zhang2021structadmm, chen2022fpc, 10103462, ghimire2023loss, guo2023sensitivity} perform during-training pruning. The SFP method \cite{he2018soft} prunes filters from scratch and omits fine-tuning procedure. This method allows pruned filters to be updated in subsequent epochs, preserving the capacity of the original network and potentially enabling pruned weights to gain importance in later epochs of training. StructADMM \cite{zhang2021structadmm} presents a framework that applies dynamic ADMM regularization during training to overcome the problem of accuracy degradation when increasing the pruning rate for both structured and unstructured pruning approaches. In the FPC method \cite{chen2022fpc}, the pruning is conducted during training. First, Singular Value Decomposition (SVD) is applied to decompose the output of the layer, and then the output feature maps with smaller singular values are removed. Consequently, the filters corresponding to those feature maps are pruned. However, this method is computationally expensive, especially in deep networks. DPU \cite{10103462} performs weight pruning during training, allowing pruned weights to be updated throughout the training process to refine the identification of candidate connections for pruning.

References \cite{han2015learning, he2019filter, lin2021filter, he2022filter, wang2023progressive} conducted post-training pruning. The approach in \cite{han2015learning} involves three key steps: first, training the network until convergence; second, pruning less important weights; and third, fine-tuning until the desired accuracy is achieved. Pruning and retraining are conducted iteratively, with weights smaller than a certain threshold being removed at each stage, followed by retraining the sparse network. These steps are repeated until all weights with minimal contribution are pruned. FilterSketch \cite{lin2021filter} proposes a pruning method for pretrained networks that preserves the second-order information, ensuring that critical information from the pretrained model is maintained during the pruning process. By encoding second-order information, this method allows a pruned network to recover its representation capacity with fine-tuning. PLFP \cite{wang2023progressive} prunes filters after training, ensuring the model's representation ability is preserved by carefully reducing filter weights based on local geometric properties, thus maintaining the original filter distribution and accuracy.

In the during-training pruning, candidate weights for pruning are identified in a non-converged model, where weights are still being updated in subsequent epochs. This approach may lead to the pruning of weights that could have become significant after further updates. We think that the importance of weights can be assessed more precisely after the model convergence. Consequently, in this work, we perform post-training pruning, in which the unpruned network is first trained until the model converges. Afterward, the weights are pruned, and the remaining weights are updated through retraining. This approach ensures that our proposed method is also usable for pruning pretrained models.

\subsection{Iterative and one-shot pruning}
Pruning methods can be divided into one-shot and iterative pruning. In iterative pruning, the pruning rate, the number of pruned layers, or both are gradually increased during training or retraining until the desired compression ratio is achieved. In contrast, one-shot pruning removes all less important weights in a single step without gradual pruning. Reference \cite{yang2022structured} conducts one-shot filter pruning, and References \cite{han2015learning, wang2023progressive, plochaet2023hardware, ghimire2023loss, chang2023iterative} perform iterative pruning. 

PLFP \cite{wang2023progressive} does not immediately drop the candidate filters; instead, it gradually decreases their impact by decreasing their values in the subsequent epochs. Moreover, incorrect candidate weights for pruning can be recovered in the next epochs. Loss-Aware Automatic Structured Pruning (LAASP) \cite{ghimire2023loss} employs an iterative filter pruning approach. This method suggests a loss-aware pruning strategy. Progressive Fisher-Rao Channel Pruning (PFRCP) \cite{xiang2026fisher} adopts a progressive strategy that alternates Fisher re-estimation and structured channel removal with short recovery fine-tuning to mitigate accuracy degradation. Structural Decoupling-Driven Loss-Aware filter pruning (SDDLA) \cite{yang2026structural} employs a two-level iterative strategy that sequentially prunes layers and progressively removes small groups of filters within each layer.

The SFP \cite{he2018soft} and Meta-attribute-based Filter Pruning
(MFP) \cite{he2022filter} methods prune filters in a one-shot manner during each training epoch. However, the pruned filters are allowed to be updated during subsequent training iterations and may therefore become active again in later epochs. As a result, these methods perform pruning in an iterative manner throughout the training process.

One-shot pruning utilizes the pruning procedure only once, making it simpler and more computationally efficient compared to iterative pruning, which repeatedly applies the pruning process over multiple stages. While iterative pruning can more effectively mitigate accuracy degradation by carefully pruning weights in stages, it has a greedy nature that might result in the removal of important weights, potentially reducing accuracy. In this work, we adopt one-shot pruning and address any accuracy degradation through retraining and updating the remaining weights.
\begin{figure*}
\centering
\includegraphics[scale=1.9]{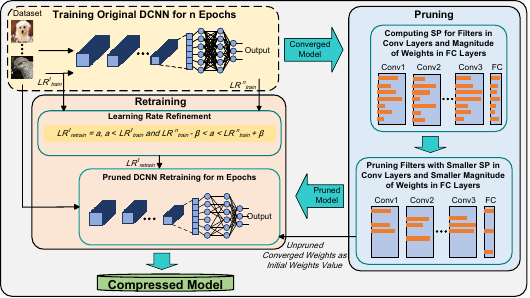}
\caption{The pruning process in SPRUNN (${LR^1}_{train}$: initial learning rate in the training, ${LR^n}_{train}$: final learning rate in the training, and ${LR^1}_{retrain}$: initial learning rate in the retraining).}
\label{fig:method}
\end{figure*}

In summary, this paper introduces SP, a novel similarity-based criterion that effectively identifies less important filters by comparing each element within a filter to the corresponding elements across all other filters in the same layer. Unlike most existing filter pruning methods, the proposed approach effectively integrates coarse-grained structured filter pruning in convolutional layers with fine-grained weight pruning in FC layers. This combination enables more precise removal of less significant weights in DNNs that include FC layers. Pruning is performed in a post-training, one-shot manner after model convergence, allowing more accurate selection of pruning candidates and eliminating the need for iterative training and pruning cycles. Moreover, it enables direct application of the method to pretrained models, thereby reducing computational overhead and facilitating faster deployment in practical applications. Following pruning, the model is retrained to update the remaining weights. To further improve accuracy, we propose a learning rate refinement strategy to be used during retraining.

\section{Methodology}\label{sec_method}
In this section, our pruning method is explained in detail. The main steps of SPRUNN are illustrated in Fig.~\ref{steps}. The operations in each step of SPRUNN are summarized as follows:
\begin{enumerate}
    \item \textbf{Original network training:} The unpruned network is trained to ensure that the weights converge to their optimal values. Since SPRUNN performs post-training pruning, this step is omitted when a pretrained model is given to SPRUNN.
    \item \textbf{Pruning criteria calculation:} SP is computed on a layer-wise basis for all filters in the convolutional layers of the converged model. For the FC layers, the magnitude of weights is considered to determine their importance.
    \item \textbf{Pruning:} In each convolutional layer, $p_{\mathrm{Conv}}\%$ of the filters with the smallest SP values are pruned, while in the FC layers, $p_{\mathrm{FC}}\%$ of the weights with the smallest magnitudes are pruned.    
    \item \textbf{Retraining and learning rate refinement:} The pruned model is retrained to adjust the unpruned weights, ensuring their compatibility with the changes and alleviating any loss in inference accuracy resulting from the pruning process. Before retraining, the learning rate is adjusted to effectively manage exploration and exploitation, thereby enhancing the test accuracy of the pruned model.
\end{enumerate}

Fig.~\ref{fig:method} illustrates the processes in various steps of SPRUNN. The details of each step are thoroughly described in the following.

\subsection{Structural proximity for filter pruning}\label{subsec_SPFP}
The proposed method introduces a novel similarity-based criterion, SP, for filter pruning in convolutional layers. As illusterated in Fig.~\ref{fig:method}, filters with smaller SP values are identified as less important and pruned from each convolutional layer. The SP criterion evaluates filter similarity by comparing each filter with all others in the same layer. In convolutional layers, where weight-sharing allows all filters to convolve with all input feature maps, removing highly similar filters has minimal impact on accuracy. This is because the remaining diverse filters can sufficiently compensate for the functionality of those that are pruned.

The SP criterion is inspired by the Crowding Distance concept from the NSGA-II algorithm \cite{996017}, a widely used multi-objective evolutionary algorithm. In NSGA-II, Crowding Distance helps maintain population diversity by measuring the distance of each individual to its nearest neighbors across all objective dimensions. This mechanism reduces the chance of selecting individuals that are too similar to others, thus enhancing population diversity. Similarly, in our approach, the SP criterion evaluates filter similarity to identify and eliminate redundant filters, thereby improving the pruning process. SP precisely computes the similarity degree of each filter by comparing it with all other filters. For this purpose, filters in each layer are treated as vectors and sorted according to each element, as shown in Equation \eqref{eq:sort}. Notably, in the following, vector refers to a filter.
\begin{equation}
\label{eq:sort}
\mathrm{sort}(\{\mathcal{F}_i\}_{i=1}^n) \; \text{by}\; \mathcal{F}_i[j] \; \forall j\in\{1,...,m\}
\end{equation}
In Equation \eqref{eq:sort}, $\{\mathcal{F}_i\}_{i=1}^n$ represents the set of $n$ filters in a layer. The notation $\mathcal{F}_i[j]$ denotes the $j$-th element of the $i$-th filter, and $m$ represents the number of elements in a filter, where $m=c\times k_h\times k_w$. Here, $c$, $k_h$, and $k_w$ indicate the number of channels, the kernel height, and the kernel width in a layer, respectively. This sorting process determines the position of each vector by evaluating its elements relative to all other vectors within the filter set. After sorting the filters, SP is computed for each filter as: 
\begin{equation}
\label{eq:SP2}
\begin{split}
SP(\mathcal{F}_i) = \sum_{j=1}^m(\mid{\mathcal{SF}[i,j]-\mathcal{SF}[i-1,j]}\mid +\\
\mid{\mathcal{SF}[i,j]-\mathcal{SF}[i+1,j]}\mid)
\end{split}
\end{equation}
where $\mathcal{SF}[i,j]$ denotes a sorted matrix of filters on all elements within the filter tensors according to Equation \eqref{eq:sort}, with $i$ indicating the filter index and $j$ representing the element index. Equation \eqref{eq:SP2} calculates the sum of distances between filter $\mathcal{F}_i$ and its two neighboring filters across all elements. This summation represents the SP of $\mathcal{F}_i$ by considering its relative similarity to adjacent filters.

SP is calculated for each filter on a layer-wise basis. Filters with lower SP values, indicating higher similarity to other filters, are identified as candidates for pruning. Since the vectors are sorted based on all their elements, and their distances to the surrounding vectors are calculated, the SP value of a filter is determined by comparing it with all other vectors. This ensures that SP offers a comprehensive evaluation of similarity within the filter set.

In Equation \eqref{eq:SP2}, $\mathcal{SF}$ is treated as a circular list. This means that after sorting the $i$-th element of vectors, the filter with the smallest value in element $i$ and the filter with the largest value in element $i$ are considered neighbors. By this arrangement, the difference between the first and last elements becomes more distinct, increasing the likelihood that filters whose elements are more frequently positioned at the extremes after sorting will remain unpruned. This is because such vectors are more distant from those whose elements are concentrated in the middle range, and therefore tend to exhibit lower similarity to other vectors.

\begin{figure}[pos={!h}]
\centering
\includegraphics[scale=0.4]{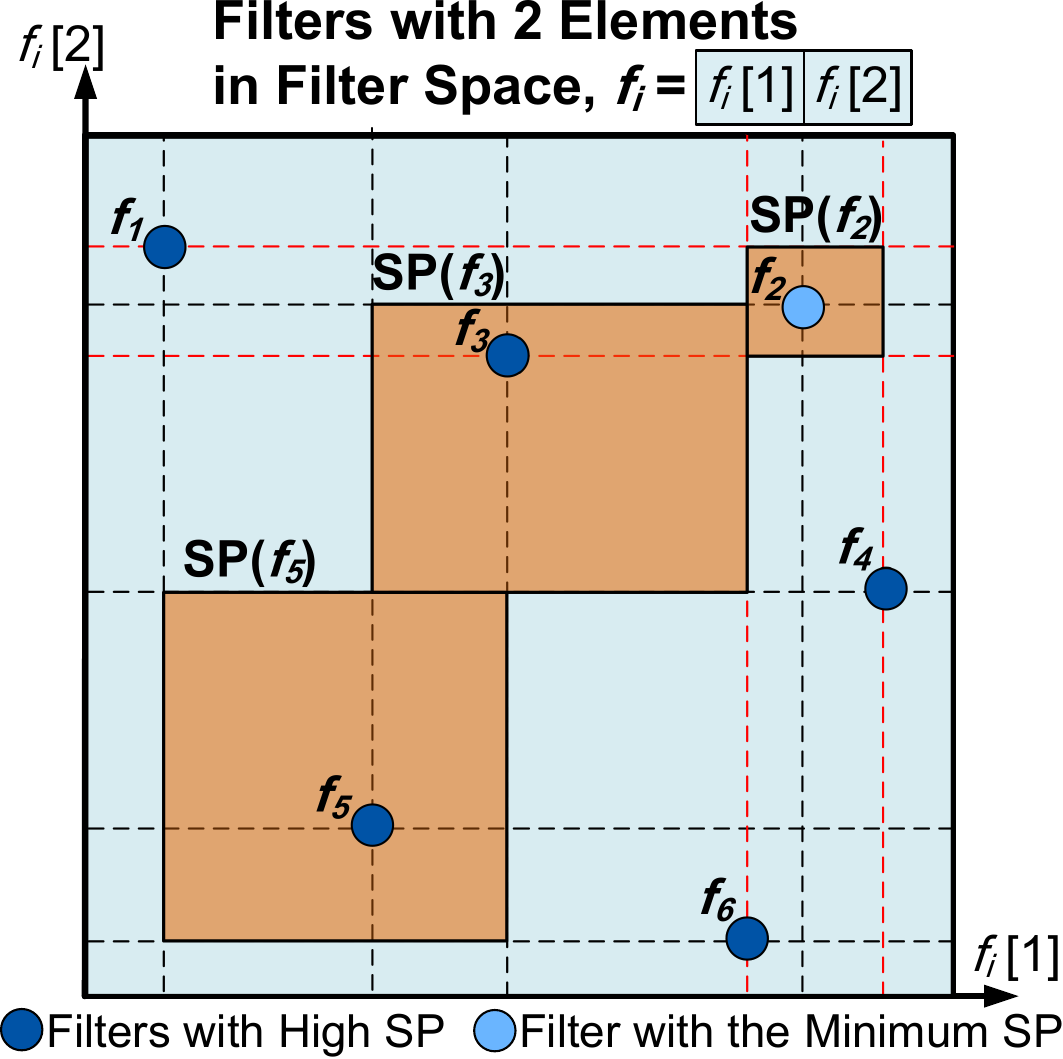}
\caption{A filter space and SP for filters with two elements.}
\label{fig:SP}
\end{figure}

Fig.~\ref{fig:SP} illustrates a schematic representation of the SP criterion. For simplicity, The filter space consists of six filters, each containing two elements. As described, the SP value for a filter is computed as the sum of its distances to the nearest filters on either side across all elements of the filter tensor. This is visually represented by the sum of the two sides of the highlighted rectangle surrounding the filter. For instance, filter $f_2$ exhibits small distances to its neighbors for both elements, resulting in a compact rectangle and, consequently, a lower SP value compared to the other filters. This makes $f_2$ a candidate for pruning. Specifically, on element ${f_i[1]}$, the distances between $f_2$ and its neighboring filters $f_4$ and $f_6$ are minimal. Similarly, on element ${f_i}[2]$, the distances to filters $f_1$ and $f_3$, located on either side of $f_2$, are also small. This suggests that after pruning, surrounding filters can cover the functionality of filter $f_2$. 

\begin{figure*}
\centering
\includegraphics[scale=1.1]{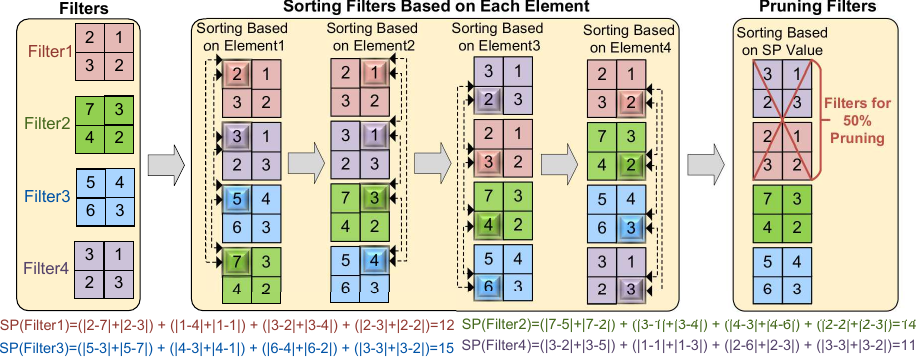}
\caption{Example of computing SP for a set of four filters, demonstrating 50\% filter pruning.}
\label{fig:SP_computation}
\end{figure*}

To illustrate how SP is calculated for a filter, Fig.~\ref{fig:SP_computation} presents an example of the SP calculation for four filters, each containing four elements. As shown the filters are sorted based on each element. Then for each filter, the difference between each element and the corresponding elements of the adjacent filters on either side is computed. Subsequently, these differences are summed up to obtain the SP value. After calculating the SP values, the filters are sorted based on the SP value, and the filters with smaller SP values are pruned. As seen in this example, filter4 and filter1 have SP values of 11 and 12, respectively, and are identified for pruning when the pruning rate is 50\%. 

SP identifies redundant filters by quantifying their similarity to neighboring filters in the sorted space using element-wise distance measurements. The aggregated distances across all elements capture how closely a filter is similar to its neighbors. Filters with lower aggregated distance values are considered as more redundant and are therefore selected for pruning, while more distinctive filters with higher distances are preserved.

\textbf{\textit{Computational Complexity Analysis:}}
Computing SP involves two main stages. In the first stage, the filters are sorted based on each element. Suppose, we have $n$ filters, each containing $m$ elements. Sorting $n$ filters on the $j$-th element has a complexity of $O(n\log{n})$. As filters include $m$ elements, the overall complexity for this stage is $O(mn\log{n})$. In the second stage, SP is computed for each filter by measuring its differences with its two neighboring filters in the element-wise sorted space and aggregating the results. The complexity of these operations for a single filter is $O(m)$, and thus $O(mn)$ for all filters. Combining the complexities of both stages, the total complexity for computing SP is: $O(mn\log{n}+mn)$. Since $nlogn$ dominates $n$ for large $n$, the complexity simplifies to $O(mn\log{n})$. 

As stated in Section \ref{subsec_SPFP} $m=c^{l}\times k_h\times k_w$, where $c^{l}$ is equal to the number of filters in layer $l-1$. For complexity analysis, we assume that the number of filters in each layer is approximately equal to $n$, hence $C^l\approx{n}$. Since $k_h$ and $k_w$ are often equal and typically small, they can be omitted in the computational complexity analysis. Therefore, the total complexity of computing SP becomes $O(n^2\log{n})$.

Notably, since SPRUNN adopts a one-shot pruning strategy, SP is computed only once for each filter during training.

\subsection{Pruning in fully connected layers}
In FC layers, unlike convolutional layers, weight-sharing is not employed, and each weight is connected to and multiplied by a single input feature. Consequently, pruning based on similarity criteria tends to be less effective. In contrast, unstructured pruning allows for more precise identification of less important weights. Therefore, we adopt a fine-grained, magnitude-based pruning strategy, in which weights with smaller magnitudes, typically those contributing least to model accuracy, are removed. Specifically, we prune $p_{\mathrm{FC}}\%$ of the weights based on their absolute values, as depicted in Fig.~\ref{fig:method}.

\subsection{Retaining and learning rate Refinement}\label{subsec:hyperparameters adjusment}
SPRUNN conducts pruning after the training and convergence of the DNN, ensuring that the optimal weight values are established and enabling the precise identification of candidates for removal. Subsequently, the pruned model is retrained to update the unpruned weights and adapt to the new structure.

Typically, the final learning rate from the training phase is used for fine-tuning \cite{renda2020comparing}. However, some previous works, such as \cite{renda2020comparing, le2021network}, have proposed techniques for selecting an optimal learning rate during retraining to achieve higher accuracy. SPRUNN suggests a learning rate refinement approach specifically for the retraining phase, as represented in Fig.~\ref{fig:method}. Notably, this refinement strategy is defined based on utilizing the final values of the remaining weights from the training phase during retraining.

In our approach, the learning rate for retraining the pruned model is set to a value smaller than its initial value during training, denoted as $\gamma^{\mathrm{initial}}_{\mathrm{retrain}}<\gamma^{\mathrm{initial}}_{\mathrm{train}}$, where $\gamma$ represents the learning rate. During training, Learning Rate Annealing is commonly applied \cite{nakamura2021learning}, where the learning rate is gradually reduced over several epochs. In this situation, the initial learning rate for retraining is determined to be close to the final learning rate of the training phase, while also satisfying the following condition:
\begin{equation}
\label{eq:lr}
\begin{aligned}
&{\gamma}^{\mathrm{final}}_{\mathrm{train}}-\beta<{\gamma}^{\mathrm{initial}}_{\mathrm{retrain}}<{\gamma}^{\mathrm{final}}_{\mathrm{train}}+\beta\\
& \text{where} 
\begin{cases}
0\leq\beta<\mid{\gamma}^{\mathrm{final}-1}_{\mathrm{train}}-{\gamma}^{\mathrm{final}}_{\mathrm{train}}\mid\\
\beta < \frac{\mid\ {\gamma}^{\mathrm{initial}}_{\mathrm{train}}-{\gamma}^{\mathrm{final}}_{\mathrm{train}}\mid}{2}
\end{cases}
\end{aligned}
\end{equation}
In Equation \eqref{eq:lr}, the parameter $\beta$ denotes a small positive value. ${\gamma}^{\mathrm{final}}_{\mathrm{train}}$ denotes the final learning rate used during training, whereas $\gamma^{\mathrm{final}-1}_{\mathrm{train}}$ refers to the learning rate value immediately preceding the final one in the training schedule.

Using learning rate refinement, we control the level of exploration to be lower than at the beginning of the unpruned network training, where the weights are randomly initialized. Conversely, since the converged weights of the unpruned network utilize as the initial weights for retraining, employing a smaller learning rate increases exploitation and is more efficient. This is because the weights are already near their optimal values, requiring only minor adjustments after pruning to reach their optimal configuration.

Algorithm~\ref{alg:pruning} provides an overview of the SPRUNN process. In this algorithm, the unpruned network ($\mathcal{W}$) is trained in line 1. The pruning process is then performed as outlined in lines 2 to 10, where $p_{\mathrm{conv}}\%$ of filers with smaller SP in the convolutional layers and $p_{\mathrm{FC}}\%$ of weights with smaller magnitude in the FC layers are pruned. Finally, lines 11 and 12 are related to the learning rate refinement and the updating of unpruned weights during retraining.
\begin{algorithm}[t]
\caption{Pruning Method.}\label{alg:pruning}
\begin{algorithmic}
\STATE \textbf{Input}: training data: $\mathcal{D}$, unpruned network: $\mathcal{W}$, number of layers: $L$, pruning rate in the conv. layers: $p_{\mathrm{conv}}$, pruning rate in the FC layers: $p_{\mathrm{FC}}$, training initial learning rate: $\gamma$.\STATE \textbf{Output}: compressed model: $\mathcal{W^*}$ where $\mathcal{W^*}<\mathcal{W}$. 
\STATE 1: Train the network to converge unpruned model
\STATE 2: For i=1 to L do
\STATE 3: \hspace{0.2cm} If i is a convolutional layer
\STATE 4: \hspace{0.5cm} Compute SP for each filter as Equation \eqref{eq:SP2}
\STATE 5: \hspace{0.5cm} Sort filters based on SP
\STATE 6: \hspace{0.5cm} Prune $p_{\mathrm{conv}}\%$ filters with smaller SP
\STATE 7: \hspace{0.2cm} Else if i is a FC layer
\STATE 8: \hspace{0.5cm} Prune $p_{\mathrm{FC}}\%$ of weights with smaller absolute \STATE \hspace{0.9cm} values 
\STATE 9: \hspace{0.2cm} End If
\STATE 10: End For
\STATE 11: Set ${\gamma^{initial}_{retrain}}=a$ where \STATE \hspace{0.6cm}($a<{\gamma^{\mathrm{initial}}_{\mathrm{train}}}$) and (${\gamma^{\mathrm{final}}_{\mathrm{train}}-\beta}<a<{\gamma^{\mathrm{final}}_{\mathrm{train}}+\beta}$)
\STATE 12: Updating unpruned weights in retraining
\end{algorithmic}
\end{algorithm}
\section{Experimental results}\label{sec_results}
In this section, we discuss the experiments conducted to evaluate the proposed method. We use various DCNNs across multiple datasets, including ResNet-20, VGG-16, and MobileNetV2 on CIFAR-10; ResNet-32 on CIFAR-10 and TinyImageNet; ResNet-56 and ResNet-110 on CIFAR-10 and CIFAR-100; ResNet-18 on CUB-200-2011; and LeNet-5 on the MNIST dataset.

\subsection{Networks and datasets}
Below is a summary of the architecture and key specifications of the DCNNs used in our experiments: 
\begin{itemize}
    \item \textbf{ResNet:} ResNet-n \cite{he2016deep} is a DCNN with $n-1$ convolutional layers and one FC layer at the end of the network. The inclusion of residual blocks in the ResNet architecture significantly improves model accuracy     
    \item \textbf{MobileNetV2:} In MobileNetV2 \cite{sandler2018mobilenetv2}, shortcut connections are used similar to ResNet architecture. It consists of multiple inverted residual blocks. Another notable feature of MobileNetV2 is the use of Depthwise Separable (DS) convolution instead of regular convolution. This approach efficiently reduces the number of multiply-accumulate (MAC) operations.
    \item \textbf{LeNet-5:} LeNet-5 \cite{lecun1998gradient} comprises 5 layers, with two convolutional layers and three FC layers.
    \item \textbf{VGG-16:} VGG-16 \cite{simonyan2014very} includes 16 layers, with 3 FC layers following 13 convolutional layers.
\end{itemize}

Below is an overview of the datasets utilized in our experiments:
\begin{itemize}
    \item \textbf{CIFAR-10:} CIFAR-10 \cite{krizhevsky2009learning} includes 50000 train samples and 10000 test samples, categorized into 10 classes. In our experiments, we employ data augmentation techniques in \cite{lee2015deeply} for the training set. No data augmentation is applied to the test set. All train and test images are normalized using $\mathrm{mean}=(0.4915,\allowbreak 0.4822,\allowbreak 0.4466)$ and
$\mathrm{std}=(0.2463,\allowbreak 0.2428,\allowbreak 0.2607)$.
    \item \textbf{CIFAR-100:} CIFAR-100 \cite{krizhevsky2009learning} is similar to CIFAR-10 but includes 100 classes, each comprising 600 images. We apply data augmentation and normalization to the training samples, following the same procedure as used for CIFAR-10.
    \item \textbf{TinyImageNet:} TinyImageNet \cite{TinyImageNet} contains 100000 training images and 10000 validation images from 200 classes selected from ImageNet \cite{5206848}. For the training set, each image is first padded with 8 pixels on all sides, after which a random crop of size 64$\times$64 is extracted and random horizontal flipping is applied. No data augmentation is used for the validation set. All training and validation images are normalized using $\mathrm{mean}=(0.485,\allowbreak 0.456,\allowbreak 0.406)$ and $\mathrm{std}=(0.229,\allowbreak 0.224,\allowbreak 0.225)$ \cite{lian2025daar}.
    \item \textbf{CUB-200-2011:} CUB-200-2011 \cite{WahCUB_200_2011} is a fine-grained dataset including 11788 high resolution bird images from 200 species, where approximately 6000 images are used for training and the remaining images are used for testing. During training, data augmentation is applied by first resizing the images to $256\times256$, followed by a random crop of size $224\times224$, random horizontal flipping, and the application of $\mathrm{ColorJitter}(\mathrm{brightness}=0.2,$ $\mathrm{contrast}=0.2,$ $\mathrm{saturation}=0.2,$ $\mathrm{hue}=0.1)$. For the test set, the images are resized to $256\times256$ and center cropped to $224\times224$. Finally, all training and test samples are normalized using the same mean and standard deviation values as those used for TinyImageNet.
    \item \textbf{MNIST:} MNIST \cite{lecun1998gradient} consists of gray-scale images of handwritten digits, with 10 classes representing digits 0 to 9. This dataset contains 60000 training samples and 10000 test samples. For training, we pad all samples with 4 pixels and then select a random crop of size 32$\times$32. No augmentation is applied to the test samples. Moreover, all train and test samples are normalized using $\mathrm{mean}=(0.1307)$ and $\mathrm{std}=(0.3081)$ \cite{Nouman}.  
\end{itemize}
\subsection{Training setups}
In the training process, the Stochastic Gradient Descent (SGD) optimizer with a Momentum of 0.9 is applied. We utilize the Cross-Entropy loss function, which is combined with the L2 regularization. 

For ResNet-18 on CUB-200-2011, transfer learning is performed using a model pretrained on the ImageNet dataset, which is then fine-tuned on CUB-200-2011. The final classification layer with 1000 output classes is replaced by a classifier with 200 output classes. Momentum is not used during fine-tuning. 
\subsubsection{Hyperparameters}
\begin{figure}[pos={!t}]
\centering
\includegraphics[scale=0.54]{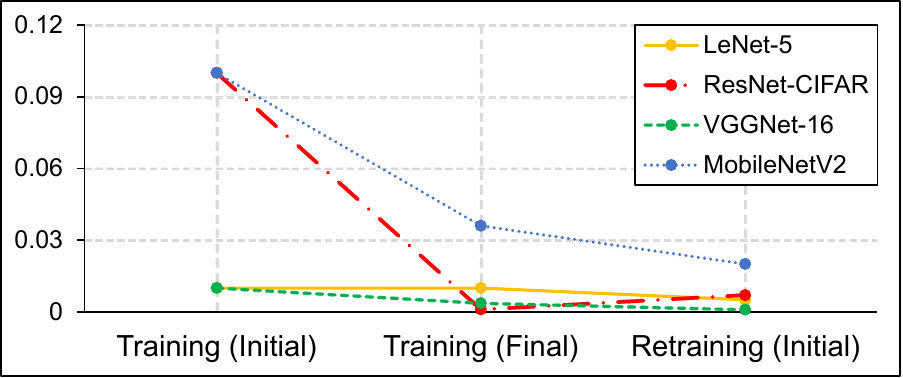}
\caption{Initial and final learning rates during training, along with the initial learning rate for retraining.}
\label{fig:paramsValues}
\end{figure}

As described in Section~\ref{subsec:hyperparameters adjusment}, we propose different values for the learning rate during training and retraining. Fig.~\ref{fig:paramsValues} represents initial and final learning rates during training, as well as the initial learning rate for retraining. As shown, the initial learning rate for retraining is lower than the initial training learning rate for all networks and remains close to the final value from training. The values of hyperparameters for various networks during training are outlined below:
\begin{itemize}
    \item \textbf{ResNet:} For training ResNet-20, ResNet-32, ResNet-56, and ResNet-110, the batch size is set to 128. The initial learning rate during training the unpruned network is 0.1, which is reduced by a factor of 0.1 at epochs 180 and 340. The regularization parameter is $3\times10^{-4}$. The initial learning rate for the retraining phase is set to $7\times10^{-3}$.
    
    For ResNet-18, the batch size is set to 32. During fine-tuning of the ImageNet-pretrained model on the CUB-200-2011 dataset, the initial learning rate is set to $4\times10^{-3}$ and reduced by a factor of 0.9 at epoch 70. The regularization parameter is set to $10^{-4}$. For retraining, the initial learning rate is adjusted to $3.7\times10^{-3}$.
    \item \textbf{MobileNetV2:} The batch size is 128. The unpruned network is trained with an initial learning rate of 0.1, which decays by a factor of 0.6 at epochs 150 and 225. The regularization parameter is set to $7\times10^{-5}$. The initial learning rate used for retraining is 0.02.
    \item \textbf{LeNet-5:} The batch size is set to 128. During training of the unpruned network, the initial learning rate is 0.01, and in retraining, it is set to $5\times10^{-3}$. The regularization parameter is $8\times10^{-5}$.
    \item \textbf{VGG-16:} The batch size is 64. The unpruned network is trained with an initial learning rate of 0.01, which is decayed by a factor of 0.6 every 120 epochs. The regularization parameter is $8\times10^{-4}$. For retraining the pruned model, the initial learning rate is set to $9\times10^{-4}$.
\end{itemize}

\subsubsection{Pruning rate}
Selecting different pruning rates for each layer is highly empirical and computationally expensive \cite{lian2025daar}. Although increasing the pruning rate generally improves the compression ratio, it may also lead to a reduction in model accuracy. Consequently, determining appropriate pruning rates requires a trade-off between compression and accuracy. To simplify this process, we employ a fixed pruning rate for all convolutional layers ($p_{\mathrm{conv}}\%$) and a separate fixed pruning rate for the FC layers ($p_{\mathrm{FC}}\%$), thereby eliminating the need for a computationally expensive layer-wise search for optimal pruning rates.

In this work, the pruning rates for convolutional layers are selected based on commonly adopted settings in previous studies, particularly those used by FPGM \cite{he2019filter}. For convolutional layers, we begin with a pruning rate of 30\% and further evaluate pruning rates of 40\%, 50\% and 60\% to assess the behavior of the proposed method under increasing levels of compression. For ResNet-18, ResNet-20, and ResNet-32, pruning rates of 30\%, 40\%, and 50\% are evaluated. For deeper architectures with larger number of parameters, such as ResNet-56, ResNet-110, and MobileNetV2, we additionally evaluate a pruning rate of 60\%.

For LeNet-5 and VGG-16, which contain FC layers that comprise a large proportion of the model parameters, different pruning rates are considered for the FC layers. Since FC layers are generally less sensitive to pruning, as will be detailed in Section~\ref{subsec:analysis pruning layers}, higher pruning rates of 90\%, 95\%, and 99\% are applied to the FC layers. In all models, the final output layer is not pruned.

\subsection{Compression metrics}
In our comparisons, the proposed method is evaluated not only based on model accuracy but also on compression metrics. We compute the decrease rate of parameters and FLOPs as compression metrics. 

\subsubsection{Number of parameters}
The number of parameters in each convolutional layer depends on both the size of each filter and the number of filters. Suppose $\mathcal{V}\in\{0,1\}^n$ is a selection vector, where $n$ denotes the number of filters. An element with value $0$ indicates that the corresponding filter is pruned, while a value of $1$ indicates that the filter is not pruned. As a result, $\mathrm{Sum}(\mathcal{V}_i)$ represents the number of unpruned filters in layers $i$. The size of each filter is determined by the number of input channels and the kernel size. Since the number of input channels in layer $i$ is equal to the number of unpruned filters in the previous layer, the number of channels is given by $\mathrm{Sum}(\mathcal{V}_{i-1})$. Consequently, the total number of parameters in each convolutional layer is computed as
\begin{equation}
\label{eq:convparams}
\mathrm{Params(Conv)} = \mathrm{Sum}(\mathcal{V}_i) \mathrm{Sum}(\mathcal{V}_{i-1}) k_h k_w
\end{equation}
where $k_h$ and $k_w$ are the height and width of the kernel, respectively. A DS convolution, which is applied in MobileNetV2, consists of a depthwise convolution followed by a pointwise convolution. The depthwise convolution has $\mathrm{Sum}(\mathcal{V}_{i-1})$ single-channel filters of size $k_h \times k_w$, while the pointwise convolution uses $\mathrm{Sum}(\mathcal{V}i)$ filters with a kernel size of $1\times1$ and $\mathrm{Sum}(\mathcal{V}_{i-1})$ input channels. Therefore, the total number of parameters in a DS convolutional (DSConv) layer is given by:
\begin{equation}
\label{eq:convparams_depthwise}
\mathrm{Params(DSConv)} = \mathrm{Sum}(\mathcal{V}_{i-1}) k_h k_w + \mathrm{Sum}(\mathcal{V}_i) \mathrm{Sum}(\mathcal{V}_{i-1})
\end{equation}

In an FC layer, the number of parameters is computed as
\begin{equation}
\label{eq:fcparams}
\mathrm{Params(FC)} = (1-p) N^l N^{l-1}
\end{equation}
where $N^l$ and $N^{l-1}$ are number of neurons in layers $l$ and $l-1$, respectively, and $p$ specifies the pruning rate. 

\subsubsection{FLOPs}
For computing the FLOPs in standard convolutional layers, the number of FLOPs required for each element in the output feature map tensor must be computed. For each element of output feature map, the MAC operation is performed between a filter of size $\mathrm{Sum}(\mathcal{V}_{i-1}) k_h k_w$ and the corresponding region of the input feature map. The total number of elements in the ouput feature map is equal to $\mathrm{Sum}(\mathcal{V}_i) H_{\mathrm{out}} W_{\mathrm{out}}$, where $H_{\mathrm{out}}$ and $W_{\mathrm{out}}$ determine the height and width of the output feature map, respectively. Therefore, the total number of FLOPs for the standard convolution (SConv) is computed as follows \cite{zu2023consecutive}:
\begin{equation}
\label{eq:convFLOPs_regular}
\begin{split}
\mathrm{FLOPs(SConv)} = \mathrm{Sum}(\mathcal{V}_i) \mathrm{Sum}(\mathcal{V}_{i-1}) k_h k_w H_{\mathrm{out}} W_{\mathrm{out}}
\end{split}
\end{equation}

Calculating the FLOPs of DS convolutional layers in MobileNetV2 requires considering both the depthwise and pointwise convolutions. In the depthwise convolution, each single-channel filter is convolved with a channel of the input feature map. Therefore, the number of FLOPs required to convolve each filter with each channel of the input feature map is $k_h k_w H_{\mathrm{out}} W_{\mathrm{out}}$. The total number of FLOPs required by the depthwise convolution (DConv), considering all input channels, is computed as
\begin{equation}
\label{eq:convFLOPs_depth}
\begin{split}
\mathrm{FLOPs(DConv)} = \mathrm{Sum}(\mathcal{V}_{i-1}) k_h k_w H_{\mathrm{out}} W_{\mathrm{out}}
\end{split}
\end{equation}
In the pointwise convolution, $\mathrm{Sum}(\mathcal{V}_i)$ filters with kernel size $1\times1$ are applied to the output of the depthwise convolution, which has dimensions $\mathrm{Sum}(\mathcal{V}_{i-1}) H_{\mathrm{out}} W_{\mathrm{out}}$. Thus, the total number of FLOPs for the pointwise convolution (PConv) is computed as follows:
\begin{equation}
\label{eq:convFLOPs_point}
\begin{split}
\mathrm{FLOPs(PConv)} = \mathrm{Sum}(\mathcal{V}_i) \mathrm{Sum}(\mathcal{V}_{i-1}) H_{\mathrm{out}} W_{\mathrm{out}}
\end{split}
\end{equation}
By combining the FLOPs computed in Equations \eqref{eq:convFLOPs_depth} and \eqref{eq:convFLOPs_point}, the total number of FLOPs in the DS convolution (DSConv) is calculated as
\begin{equation}
\label{eq:convFLOPs_depthwise}
\begin{split}
\mathrm{FLOPs(DSConv)} = \mathrm{Sum}(\mathcal{V}_{i-1}) k_h k_w H_{\mathrm{out}} W_{\mathrm{out}}+\\
\mathrm{Sum}(\mathcal{V}_i) \mathrm{Sum}(\mathcal{V}_{i-1}) H_{\mathrm{out}} W_{\mathrm{out}}
\end{split}
\end{equation}

Computing FLOPs in FC layers is performed as Equation \eqref{eq:fcparams}, as each weight has a multiplication with a input feature.

It is important to note that biases are excluded from the calculations used to determine the number of parameters and FLOPs.

\subsection{Comparison with SOTA}
To illustrate the effectiveness of our method, we compare SPRUNN with the unpruned model as a baseline and with SOTA pruning methods, including Iterative Clustering Pruning (ICP) \cite{chang2023iterative}, FPC \cite{chen2022fpc}, LAASP \cite{ghimire2023loss}, Concurrent Training and Layer Pruning (CTLP) \cite{guenter2026concurrent}, Sensitivity Pruner (SP) \cite{guo2023sensitivity}, Torque \cite{gupta2024torque}, Han et al. \cite{han2015learning}, SFP \cite{he2018soft}, FPGM \cite{he2019filter}, MFP \cite{he2022filter}, DPU \cite{10103462}, Pruning with Output Error Minimization (POEM) \cite{kamma2023pruning}, TrunkPruning \cite{kim2023trunk}, Joint pruning \cite{lee2022dnn}, Similarity-based Filter
Pruning (SFP-D) \cite{li2025sfp}, Dual Attention cooperative Adaptive pruning Rate (DAAR) \cite{lian2025daar}, FilterSketch \cite{lin2021filter}, EPruner \cite{lin2021network}, EACP \cite{liu2023eacp}, Iterative Energy-Aware Channel Pruning (IEACP) \cite{pinto2026towards}, SOGP \cite{shi2024sparse}, Layer Interdependence-Aware Pruning (LIAP) \cite{tofigh2026resource}, SpectralNorm \cite{tyche2024improving}, PFRCP \cite{xiang2026fisher}, SDDLA \cite{yang2026structural}, FCS-MLFT \cite{yang2022structured}, Hessian-Aware Pruning (HAP) and HAP+IMPLANT \cite{yu2022hessian}, Finer-Grained Structured Element Pruning (FGSEP) \cite{ZHANG2026134207}, Zhang et al. \cite{ZHANG2023116}, StructADMM \cite{zhang2021structadmm}, Zheng et al. \cite{zheng2024novel}, and CLCS \cite{zu2023consecutive}. Our comparisons in Tables~\ref{tab-res-20} to ~\ref{tab-vgg-16} are based on test accuracy, parameter reduction, and FLOPs reduction. All tables are sorted in descending order of test accuracy, and the highest values for accuracy, parameter reduction, and FLOPs reduction are highlighted in bold. In these tables, "N/A" indicates that the corresponding value is not reported in the original paper. The numbers listed with the SPRUNN method denote the pruning rates. The first number represents $p_{\mathrm{conv}}$, and the second shows $p_{\mathrm{FC}}$. If only a single value is provided, it refers to the pruning rate in the convolutional layers. Notice that $p_{\mathrm{conv}}$ denotes the proportion of filters pruned in a convolutional layer, rather than the overall parameter reduction rate. As pruning filters in one layer also eliminates the corresponding input channels in the subsequent layer, the actual parameter reduction differs from the filter pruning rate.

\subsubsection{Comparisons of models with pruning only convolutional layers}
In the ResNet models and MobileNetV2, all pruned layers are convolutional layers. As seen in Table~\ref{tab-res-20}, SPRUNN achieves the highest accuracy of 91.53\% and the greatest FLOPs reduction of 61.27\% for ResNet-20, outperforming other methods. Our method also achieves more than 74\% reductions in both parameters and FLOPs while maintaining an accuracy of 90.95\%.
\begin{table}[pos={h}]
\fontsize{8pt}{8pt}\selectfont
\begin{center}
\begin{threeparttable}
\caption{Results for ResNet-20 on CIFAR-10.}
\label{tab-res-20}
\begin{tabular}{@{} llll@{} }
\toprule
Method&Acc. (\%)&Params. $\downarrow$\tnote{*} (\%)&FLOPs $\downarrow$ (\%)\\
\midrule
Baseline&92.88&0&0\\
\textbf{SPRUNN(40$\%$)}\tnote{\textdagger}&\textbf{91.53}&62.23&61.27\\
FPGM \cite{he2019filter}&91.09&N/A\tnote{\textdaggerdbl}&42.20\\
SFP-D \cite{li2025sfp}&91.06&42.24&50.19\\
DPU \cite{10103462}&91.05&N/A&43\\
\textbf{SPRUNN(50$\%$)}&90.95&\textbf{74.90}&\textbf{74.84}\\
SFP \cite{he2018soft}&90.83&N/A&42.20\\
\bottomrule
\end{tabular}
\begin{tablenotes}
   \item [*] The symbol $\downarrow$ denotes reduction.
   \item [\textdagger] The number indicates the filter pruning rate, which differs from the parameter reduction rate.
   \item [\textdaggerdbl] N/A denotes that it is not reported in the original paper. 
\end{tablenotes}
\end{threeparttable}
\end{center}
\end{table}

Table~\ref{tab-res-32} presents the results of ResNet-32 on CIFAR-10 and TinyImageNet. As shown, SPRUNN achieves the highest accuracy of 92.88\% among the compared methods on CIFAR-10 while reducing the number of parameters by more than 62\%. Although FGSEP achieves the highest parameter reduction (82.56\%), SPRUNN maintains a competitive accuracy of 92.15\% with a 50\% filter-pruning rate, while reducing FLOPs by 73.67\%, outperforming the other methods in terms of FLOPs reduction. The results on TinyImageNet further confirm the effectiveness of SPRUNN, which achieves an accuracy of 54.37\% while reducing both parameters and FLOPs by approximately 61\%, outperforming the SOGP method, which achieves an accuracy of 53.40\% with a parameter reduction of 55.25\%.
\begin{table}[pos={t}]
\fontsize{5.9pt}{5.9pt}\selectfont
\begin{center}
\caption{Results for ResNet-32 on CIFAR-10 and TinyImageNet.}
\label{tab-res-32}
\begin{tabular}{@{} lllll@{}}
\toprule
Dataset&Method&Acc. (\%)&Params. $\downarrow$ (\%)&FLOPs $\downarrow$ (\%)\\
\midrule
\multirow{10}{*}{CIFAR-10}&Baseline&93.77&0&0\\
&\textbf{SPRUNN(40$\%$)}&\textbf{92.88}&62.30&61.40\\
&LAASP \cite{ghimire2023loss}&92.64&N/A&53.30\\
&Zheng et al. \cite{zheng2024novel}&92.39&N/A&55.60\\
&FGSEP \cite{ZHANG2026134207}&92.31&\textbf{82.56}&70.02\\
&FPGM \cite{he2019filter}&92.31&N/A&41.50\\
&\textbf{SPRUNN(50$\%$)}&92.15&74.94&\textbf{73.67}\\
&SFP \cite{he2018soft}&92.08&N/A&41.50\\
&DPU \cite{10103462}&91.88&N/A&43.30\\
&MFP \cite{he2022filter}&91.85&N/A&53.20\\
\hline
\multirow{3}{*}{TinyImageNet}&Baseline&56.34&0&0\\
&\textbf{SPRUNN(40$\%$)}&\textbf{54.37}&\textbf{61.69}&\textbf{61.40}\\
&SOGP \cite{shi2024sparse}&53.40&55.25&N/A\\
\bottomrule
\end{tabular}
\end{center}
\end{table}

Table~\ref{tab-res-56} demonstrates the results of ResNet-56 on the CIFAR-10 and CIFAR-100 datasets. The results show that SPRUNN achieves an accuracy of 93.86\% with 40\% filter pruning on CIFAR-10. While the SP and FPC methods achieve slightly higher accuracies of 94.08\% and 94.01\%, respectively, our method surpasses the SP method in both parameter and FLOPs reductions, achieving 62.35\% and 61.48\%, respectively. Moreover, it significantly outperforms FPC in FLOPs reduction. With 50\% filter pruning, SPRUNN attains an accuracy of 93.12\%, while it achieves the greatest compression, with close to 75\% reduction in both parameters and FLOPs, compared to other methods while achieving the highest compression among the compared methods, with approximately 74.90\% reductions in both parameters and FLOPs. With 60\% filter pruning, our method achieves parameters and FLOPs reductions of 83.33\% and 82.50\%, respectively, while maintaining an accuracy of 92.85\%, which is acceptable relative to other approaches. The results on CIFAR-100 determine that SPRUNN achieves 70.91\% accuracy, only 0.39\% and 0.01\% lower than the FPC and ICP methods, respectively. With 50\% filter pruning, SPRUNN achieves approximately 74\% reduction in both parameters and FLOPs while maintaining 69.43\% accuracy, only a 2.80\% drop compared to the baseline.

\begin{table}[pos={t}]
\fontsize{6pt}{6pt}\selectfont
\begin{center}
\caption{Results for ResNet-56 on CIFAR-10 and CIFAR-100.}
\label{tab-res-56}
\begin{tabular}{@{} lllll@{}}
\toprule
Dataset&Method&Acc. (\%)&Params. $\downarrow$ (\%)&FLOPs $\downarrow$ (\%)\\
\midrule
\multirow{25}{*}{CIFAR-10}&Baseline&94.44&0&0\\
&SP \cite{guo2023sensitivity}&\textbf{94.08}&59.03&47.68\\
&FPC \cite{chen2022fpc}&94.01&65.88&35.71\\
&\textbf{SPRUNN(40$\%$)}&93.86&62.35&61.48\\
&PFRCP \cite{xiang2026fisher}&93.85&71&56.20\\
&FGSEP \cite{ZHANG2026134207}&93.84&78.06&70.43\\
&Torque \cite{gupta2024torque}&93.76&65&53.49\\
&MFP \cite{he2022filter}&93.56&N/A&52.60\\
&HAP+IMPLANT \cite{yu2022hessian}&93.52&65.51&62.85\\
&FPGM \cite{he2019filter}&93.49&N/A&52.60\\
&LAASP \cite{ghimire2023loss}&93.49&N/A&52.60\\
&CLCS \cite{zu2023consecutive}&93.47&N/A&54.30\\
&SFP \cite{he2018soft}&93.35&N/A&52.60\\
&SFP-D \cite{li2025sfp}&93.29&63.40&63.31\\
&FilterSketch \cite{lin2021filter}&93.19&41.20&41.50\\
&EPruner \cite{lin2021network}&93.18&54.20&61.33\\
&ICP \cite{chang2023iterative}&93.16&73.57&71.32\\
&\textbf{SPRUNN(50$\%$)}&93.12&74.97&74.91\\
&EACP \cite{liu2023eacp}&93.11&63.50&68\\
&Zheng et al. \cite{zheng2024novel}&93.04&N/A&61.50\\
&\textbf{SPRUNN(60$\%$)}&92.85&\textbf{83.33}&\textbf{82.50}\\
&DPU \cite{10103462}&92.75&N/A&43.40\\
&LIAP \cite{tofigh2026resource}&92.56&74.12&74.36\\
&Zhang et al. \cite{ZHANG2023116}&92.48&71.80&72.30\\
&CTLP \cite{guenter2026concurrent}&92.45&66.93&51.20\\
\hline
\multirow{8}{*}{CIFAR-100}&Baseline&72.23&0&0\\
&FPC \cite{chen2022fpc}&\textbf{71.30}&69.76&40.91\\
&ICP \cite{chang2023iterative}&70.92&N/A&56.20\\
&\textbf{SPRUNN(30$\%$)}&70.91&49.60&47.40\\
&CTLP \cite{guenter2026concurrent}&70.45&59.22&52.11\\
&\textbf{SPRUNN(40$\%$)}&70.17&62.19&61.48\\
&TrunkPruning \cite{kim2023trunk}&69.90&N/A&53.41\\
&\textbf{SPRUNN(50$\%$)}&69.43&\textbf{74.79}&\textbf{74.91}\\
\bottomrule
\end{tabular}
\end{center}
\end{table}

For ResNet-110, Table~\ref{tab-res-110} illustrates that SPRUNN achieves an accuracy of 94.33\% on CIFAR-10, which is only 0.03\% and 0.18\% lower compared to the ICP and FPC methods, respectively. Nevertheless, our method demonstrates a FLOPs reduction ratio of 61.53\%, surpassing both ICP at 55.05\% and FPC at 48.19\%. The PFRCP method achieves the highest parameter reduction among the methods listed in the table, removing 85.5\% of the parameters. However, with a 50\% filter pruning rate, SPRUNN achieves a higher FLOPs reduction of 74.96\%, compared with 66.8\% for PFRCP, while maintaining a higher accuracy of 94.13\%, compared with 93.95\% for PFRCP. With a filter pruning rate of 60\%, SPRUNN achieves a competitive accuracy of 93.80\% while reducing both the number of parameters and FLOPs by more than 82\%. The experiments on CIFAR-100 demonstrate that SPRUNN achieves an accuracy of 73.32\%, which surpasses the baseline, while reducing parameters and FLOPs by 49.71\% and 47.48\%, respectively. With 50\% filter pruning of ResNet-110 on CIFAR-100, SPRUNN attains an accuracy of 71.12\%, representing only a 1.65\% drop compared to the baseline, along with approximately 74.90\% reductions in both parameters and FLOPs.

\begin{table}[pos={b}]
\fontsize{6pt}{6pt}\selectfont
\begin{center}
\caption{Results for ResNet-110 on CIFAR-10 and CIFAR-100.}
\label{tab-res-110}
\begin{tabular}{@{} lllll@{}}
\toprule
Dataset&Method&Acc. (\%)&Params. $\downarrow$ (\%)&FLOPs $\downarrow$ (\%)\\
\midrule
\multirow{22}{*}{CIFAR-10}&Baseline&94.50&0&0\\
&FPC \cite{chen2022fpc}&\textbf{94.54}&67.05&48.19\\
&ICP \cite{chang2023iterative}&94.36&61.85&55.05\\
&\textbf{SPRUNN(40$\%$)}&94.33&62.38&61.53\\
&FGSEP \cite{ZHANG2026134207}&94.32&82.65&70.28\\
&LAASP \cite{ghimire2023loss}&94.17&N/A&52.50\\
&\textbf{SPRUNN(50$\%$)}&94.13&74.98&74.96\\
&Zheng et al. \cite{zheng2024novel}&94.11&N/A&61.60\\
&PFRCP \cite{xiang2026fisher}&93.95&\textbf{85.50}&66.80\\
&SDDLA \cite{yang2026structural}&93.91&69.19&69.66\\
&SFP \cite{he2018soft}&93.86&N/A&40.80\\
&\textbf{SPRUNN(60$\%$)}&93.80&83.35&\textbf{82.56}\\
&DPU \cite{10103462}&93.76&N/A&43.60\\
&CLCS \cite{zu2023consecutive}&93.74&N/A&67.30\\
&FPGM \cite{he2019filter}&93.74&N/A&52.30\\
&Zhang et al. \cite{ZHANG2023116}&93.73&68.30&71.60\\
&EPruner \cite{lin2021network}&93.62&65.91&76.30\\
&FilterSketch \cite{lin2021filter}&93.44&59.90&63.30\\
&MFP \cite{he2022filter}&93.38&N/A&62.80\\
&EACP \cite{liu2023eacp}&93.35&65.90&68.30\\
&LIAP \cite{tofigh2026resource}&93.28&69.20&69.40\\
&CTLP \cite{guenter2026concurrent}&92.83&79.44&68.63\\
\hline
\multirow{5}{*}{CIFAR-100}&Baseline&72.77&0&0\\
&\textbf{SPRUNN(30$\%$)}&\textbf{73.32}&49.71&47.48\\
&FPC \cite{chen2022fpc}&72.39&73.98&51.67\\
&\textbf{SPRUNN(40$\%$)}&72.28&62.30&61.53\\
&\textbf{SPRUNN(50$\%$)}&71.12&\textbf{74.90}&\textbf{74.96}\\
\bottomrule
\end{tabular}
\end{center}
\end{table}

Table~\ref{tab-res-18} presents results of ResNet-18 on the CUB-200-2011 dataset. SPRUNN achieves an accuracy of 71.63\%, which is 0.23\% higher than that of the POEM method, while removing 74.75\% of the parameters. In addition, SPRUNN reduces FLOPs by 73.3\%, which is only 0.03\% lower than the reduction achieved by POEM. Although CUB-200-2011 is a challenging fine-grained dataset with subtle inter-class differences, SPRUNN achieves significant accuracy even at high parameters and FLOPs pruning rates.

\begin{table}[pos={!h}]
\fontsize{9pt}{9pt}\selectfont
\begin{center}
\caption{Results for ResNet-18 on CUB-200-2011.}
\label{tab-res-18}
\begin{tabular}{@{} llll@{}}
\toprule
Method&Acc. (\%)&Params. $\downarrow$ (\%)&FLOPs $\downarrow$ (\%)\\
\midrule
Baseline&77.13&0&0\\
\textbf{SPRUNN(50$\%$)}&\textbf{71.63}&\textbf{74.75}&73.37\\
POEM \cite{kamma2023pruning}&71.40&N/A&\textbf{73.40}\\
\bottomrule
\end{tabular}
\end{center}
\end{table}

Table~\ref{tab-mobilenet} confirms the efficiency of our method in achieving high compression on MobileNetV2 while maintaining acceptable accuracy. SPRUNN attains an accuracy of 94.76\%, slightly lower than 95.30\% achieved by StructADMM. However, SPRUNN demonstrates a decrease rate of approximately 83\% in both parameters and FLOPs, surpassing all other methods.

\begin{table}[pos={!t}]
\fontsize{8pt}{8pt}\selectfont
\begin{center}
\caption{Results for MobileNetV2 on CIFAR-10.}
\label{tab-mobilenet}
\begin{tabular}{@{} llll@{}}
\toprule
Method&Acc. (\%)&Params. $\downarrow$ (\%)&FLOPs $\downarrow$ (\%)\\
\midrule
Baseline&95.07&0&0\\
StructADMM \cite{zhang2021structadmm}&\textbf{95.30}&N/A&69.70\\
\textbf{SPRUNN(60$\%$)}&94.76&\textbf{83.67}&\textbf{83.45}\\
SFP-D \cite{li2025sfp}&94.06&43.07&48.66\\
FCS-MLFT \cite{yang2022structured}&93.03&71.75&N/A\\
Zheng et al. \cite{zheng2024novel}&92.57&N/A&62.30\\
\bottomrule 
\end{tabular}
\end{center}
\end{table}
\begin{figure}[pos={!b}]
\centering
\subfloat[ResNet-110 on CIFAR-10.]
{\includegraphics[scale=0.46]{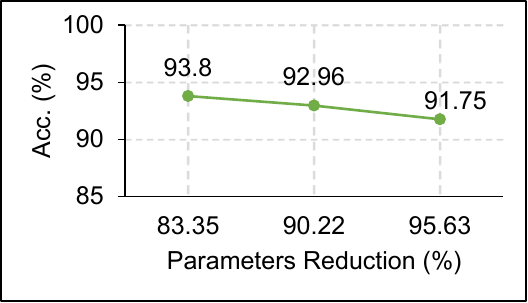}\label{fig:increasepruningrateres}}
\hfill
\subfloat[MobileNetV2 on CIFAR-10.]
{\includegraphics[scale=0.46]{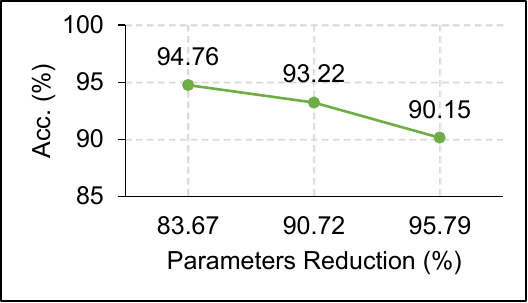}\label{fig:increasepruningratemobile}}
\caption{Changes in test accuracy as the parameter reduction increases in SPRUNN.}
\label{fig:increasepruningrate}
\end{figure}

Fig.~\ref{fig:increasepruningrate} demonstrates how test accuracy varies with increasing pruning rates using SPRUNN on ResNet-110 and MobileNetV2, evaluated on the CIFAR-10 dataset. As illustrated in Fig.~\ref{fig:increasepruningrateres}, ResNet-110 achieves an accuracy of 93.80\% after a parameter reduction of 83.35\%. When the reduction increases to 90.22\%, the model still attains a notable accuracy of 92.96\%. Even with a 95.63\% reduction in parameters, SPRUNN maintains robust performance, reaching 91.75\% accuracy. For MobileNetV2 in Similarly, Fig.~\ref{fig:increasepruningratemobile} shows that SPRUNN achieves 94.76\% accuracy after an 83.67\% parameter reduction. At a 90.72\% reduction, the model retains a high accuracy of 93.22\%, and even with a 95.79\% reduction, it reaches the accuracy of 90.15\%. These results confirm the robustness of SPRUNN under high pruning rates.

\subsubsection{Comparisons of models including convolutional and FC layers for pruning}
We evaluate the SPRUNN method on DCNNs, including FC layers in the pruning process alongside convolutional layers, to examine the effectiveness of combining the proposed coarse-grained filter pruning for convolutional layers with fine-grained pruning for FC layers.

Table~\ref{tab-lenet} presents the results for LeNet-5. SPRUNN achieves an accuracy of 99.24\%, outperforming other methods, while reducing the number of parameters by 91.78\%.
\begin{table}[pos={!b}]
\fontsize{7pt}{7pt}\selectfont
\begin{center}
\begin{threeparttable}
\caption{Results for LeNet-5 on MNIST.}
\label{tab-lenet}
\begin{tabular}{@{} llll@{}}
\toprule
Method&Acc. (\%)&Params. $\downarrow$ (\%)&FLOPs $\downarrow$ (\%)\\
\midrule
Baseline&99.47&0&0\\
\textbf{SPRUNN(40$\%$,95$\%$)}\tnote{*}&\textbf{99.24}&91.78&56.27\\
Han et al. \cite{han2015learning}&99.23&91.65&84\\
Joint pruning \cite{lee2022dnn}&99.20&98.70&\textbf{94.10}\\
FGSEP \cite{ZHANG2026134207}&99.20&\textbf{99.90}&90.60\\
SOGP \cite{shi2024sparse}&99.18&98.96&91.14\\
\bottomrule 
\end{tabular}
\begin{tablenotes}
   \item [*] The first and second numbers denote $p_{\mathrm{conv}}\%$ and $p_{\mathrm{FC}}\%$, respectively.
\end{tablenotes}
\end{threeparttable}
\end{center}
\end{table}

The results for VGG-16 presented in Table~\ref{tab-vgg-16} illustrate that SPRUNN achieves an accuracy of 94.25\%, which is only 0.08\% lower than the CLCS method. However, our method reduces FLOPs by 57.02\%, surpassing CLCS, which achieves a 52.20\% reduction. With 50\% filter pruning in the convolutional layers and 99\% weight pruning in the FC layers, SPRUNN attains the highest parameter reduction of 96.33\%, matching that obtained by the ICP method. Additionally, our method outperforms ICP in test accuracy, reaching 93.40\% compared to 93.29\%.
\begin{table}
\fontsize{7pt}{7pt}\selectfont
\begin{center}
\caption{Results for VGG-16 on CIFAR-10.}
\label{tab-vgg-16}
\begin{tabular}{@{} llll@{}}
\toprule
Method&Acc. (\%)&Params. $\downarrow$ (\%)&FLOPs $\downarrow$ (\%)\\
\midrule
Baseline&94.50&0&0\\
CLCS \cite{zu2023consecutive}&\textbf{94.33}&N/A&52.20\\
\textbf{SPRUNN(40$\%$, 40$\%$)}&94.25&42.59&57.02\\
FPC \cite{chen2022fpc}&94.08&95.17&65.62\\
SpectralNorm \cite{tyche2024improving}&94.05&83.20&53.60\\
SDDLA \cite{yang2026structural}&93.88&89.12&57.30\\
LAASP \cite{ghimire2023loss}&93.79&N/A&60.50\\
Zhang et al. \cite{ZHANG2023116}&93.76&80.80&73.70\\
DAAR \cite{lian2025daar}&93.75&84.30&84.10\\
FGSEP \cite{ZHANG2026134207}&93.66&94.37&73.73\\
Zheng et al. \cite{zheng2024novel}&93.48&N/A&63.50\\
\textbf{SPRUNN(50$\%$, 99$\%$)}&93.40&\textbf{96.33}&83.56\\
PFRCP \cite{xiang2026fisher}&93.39&90.20&50.60\\
HAP \cite{yu2022hessian}&93.37&94.90&79.70\\
FCS-MLFT \cite{yang2022structured}&93.31&91.77&59.04\\
ICP \cite{chang2023iterative}&93.29&\textbf{96.33}&86.20\\
SFP-D \cite{li2025sfp}&93.26&92.12&61.30\\
EACP \cite{liu2023eacp}&93.20&77.20&77.40\\
StructADMM \cite{zhang2021structadmm}&93.20&92.70&67.74\\
EPruner \cite{lin2021network}&93.08&76.34&\textbf{88.80}\\
SP \cite{guo2023sensitivity}&93.05&80.43&48\\
CTLP \cite{guenter2026concurrent}&92.74&91.55&63.22\\
IEACP \cite{pinto2026towards}&92.41&89.67&67.41\\
\bottomrule
\end{tabular}
\end{center}
\end{table}

The results demonstrate that the proposed SP criterion effectively identifies less important filters, enabling SPRUNN to achieve a effective trade-off between compression and accuracy across diverse network architectures. In convolutional architectures such as ResNet and MobileNetV2, SPRUNN attains substantial reductions in parameters and FLOPs while maintaining high accuracy. In networks containing both convolutional and FC layers, such as LeNet-5 and VGG-16, the combination of coarse-grained filter pruning and fine-grained weight pruning achieves significant parameter reduction with minimal accuracy degradation.
\subsection{Structural proximity efficiency}
\begin{figure}[pos={!h}]
\centering
\includegraphics[scale=0.65]{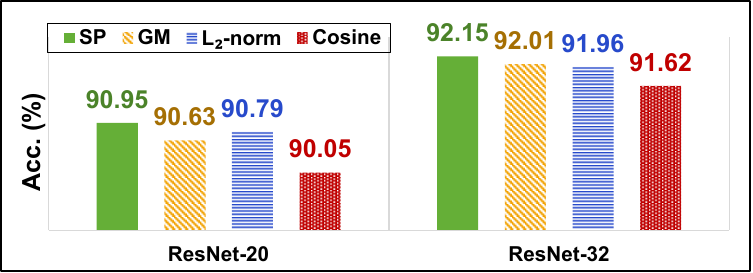}
\caption{Test accuracy comparison of pruned models obtained using the proposed SP criterion, GM, Cosine and $\mathrm{L_2-norm}$ criteria.}
\label{fig:simcriterion}
\end{figure}

To demonstrate the effectiveness of the proposed SP criterion for filter pruning, we compare it with two commonly used similarity-based pruning criteria: GM and Cosine, and the commonly used magnitude-based pruning criterion: $\mathrm{L_2-norm}$. The evaluation is conducted on ResNet-20 and ResNet-32 architectures with 50\% filter pruning. For a fair comparison, all other parameters and conditions of the pruning algorithm are kept identical across experiments. As illustrated in Fig.~\ref{fig:simcriterion}, the SP criterion yields the highest accuracy across both evaluated models, indicating its superior ability to maintain model performance after pruning. This advantage comes from the ability of SP to identify similar filters by evaluating their element-wise distances to all other filters within a layer, enabling more informed and structurally aware pruning decisions.
\subsection{Learning rate refinement efficiency}\label{sec:hyperparameter}
\begin{figure}[pos={!h}]
\centering
\includegraphics[scale=0.54]{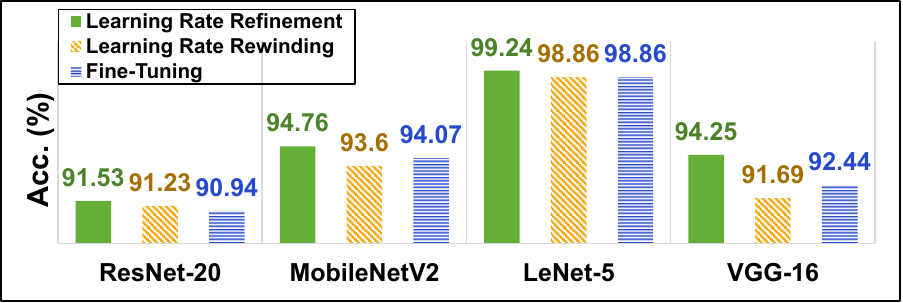}
\caption{Test accuracy comparison of pruned models retrained using learning rate refinement, learning rate rewinding, and fine-tuning.}
\label{fig:params}
\end{figure}

In this section, we evaluate the efficiency of the proposed learning rate refinement strategy during the retraining phase of the pruned model. To this purpose, we compare our approach with two commonly used strategies:
\begin{enumerate}
    \item \textbf{Fine-tuning:} In this strategy, the final learning rate value from the original training process is used for retraining.
    \item \textbf{Learning rate rewinding:} In this strategy, a learning rate from an earlier epoch, prior to the final epoch, of the original training is utilized for retraining \cite{renda2020comparing}.
\end{enumerate}
The comparison is conducted on ResNet-20, MobileNetV2, LeNet-5, and VGG-16 models, with all other training parameters kept constant throughout the experiments. The results illustrated in Fig.~\ref{fig:params} confirm that our learning rate refinement approach significantly enhances the accuracy of the pruned model compared to the other two approaches.

\begin{figure}[pos={!h}]
\centering
\includegraphics[scale=0.73]{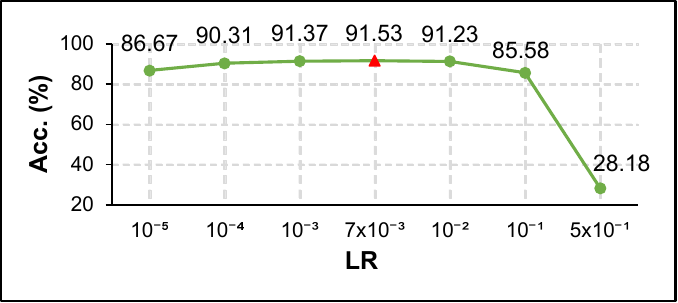}
\caption{Effect of the retraining learning rate on the test accuracy of the pruned ResNet-20 model.}
\label{fig:diff-lr}
\end{figure}

In another experiment, we evaluate the effect of different Learning Rate (LRs) during retraining while keeping all other training settings fixed, as shown in Fig.~\ref{fig:diff-lr}. The experiment is conducted on ResNet-20. The highest accuracy, 91.53\%, is achieved with a learning rate of $7\times10^{-3}$, which is selected by the proposed learning rate refinement strategy. The results show that accuracy decreases as the learning rate deviates from this value. In particular, using the initial training learning rate (0.1) reduces the accuracy to 85.58\%, while excessively large and small learning rates, such as 0.5 and $10^{-5}$, decrease the accuracy to 28.18\% and 86.67\%, respectively.

These results confirms that selecting an appropriate learning rate during retraining is critical for achieving high accuracy. Since the remaining weights are already close to their optimal values after the initial training phase, the weights require only minor updates during retraining to adapt to the modified structure. Excessively large learning rates lead to large steps in the optimization algorithm, causing the weights to deviate significantly from their optimal values. In contrast, excessively small learning rates limit the exploration capability of the optimization process, preventing the weights from efficiently adapting to the pruned architecture.

\subsection{Analysis of pruning in convolutional and FC layers}\label{subsec:analysis pruning layers}
In this section, the effect of pruning filters in the convolutional layers and weights in the FC layers on accuracy and compression metrics is investigated. Experiments are conducted on LeNet-5 and VGG-16. Table \ref{tab-LeNet-VGG-Info} presents the distributions of parameters and FLOPs across convolutional and FC layers. As depicted, most parameters reside in the FC layers, whereas most FLOPs are concentrated in the convolutional layers. In LeNet-5 more than 95\% of the parameters are in the three FC layers, while over 85\% of the FLOPs are in the two convolutional layers. In VGG-16, approximately 89\% of parameters exist in 3 FC layers, whereas only about 11\% are in the 13 convolutional layers. Conversely, more than 72\% of the FLOPs are in the convolutional layers, with the remaining in the FC layers. As a result, a high pruning rate of parameters in the convolutional layers is effective in reducing FLOPs, while a high pruning rate in the FC layers can significantly decrease the total number of parameters in the pruned model.
\begin{table}
[width=.9\linewidth,cols=4,pos=t]
\begin{center}
\caption{Ratio of parameters and FLOPs in the convolutional and FC Layers in LeNet-5 and VGG-16.}
\label{tab-LeNet-VGG-Info}
\begin{tabular*}{\tblwidth}{@{} LLLLLL@{}}
\toprule
\multirow{2}{*}{Model}&\multicolumn{2}{L}{Parameters(\%)}&\multicolumn{2}{L}{FLOPs(\%)}\\\cline{2-5}
&Conv&FC&Conv&FC\\
\midrule
LeNet-5&4.17&95.83&85.85&14.15\\
\hline
VGG-16&10.96&89.04&72.37&27.63\\
\bottomrule
\end{tabular*}
\end{center}
\end{table}

Table \ref{tab-Conv-FC} compares the accuracy, parameter reduction, and FLOPs reduction of LeNet-5 and VGG-16 with different pruning rates for convolutional and FC layers. For LeNet-5, with 40\% filter pruning in the convolutional layers and 40\% weight pruning applied to the FC layers, the model achieves the parameters and FLOPs reduction of 40.01\% and 48.60\%, respectively. When the filter pruning rate of the convolutional layers remains at 40\% but the pruning rate of the FC layers increases to 95\%, the parameter reduction rises to 91.78\%, and the FLOPs reduction increases to 56.78\%. Conversely, if the pruning rate of the FC layers is kept constant while increasing the pruning rate of filters in the convolutional layers to 95\%, the impact on FLOPs is more significant than on parameters. In this scenario, FLOPs are reduced by 86.13\%, while the parameter reduction is 41.73\%.

In another experiment on VGG-16, we first prune 50\% of filters in the convolutional layers, along with 50\% weight pruning in the FC layers. The results show a 52.72\% reduction in parameters and a 67.99\% reduction in FLOPs. Next, we keep the pruning rate of filters in the convolutional layers at 50\% and increase the pruning rate of the FC layers to 90\%. This change leads to an approximately 35\% greater reduction in parameters compared to the approximately 11\% reduction in FLOPs. Then, we reverse the pruning rates by increasing the pruning rate of filters in the convolutional layers to 90\% while keeping the pruning rate of the FC layers at 50\%. In this case, the reduction in parameters increases by only 2.63\% while the reduction in FLOPs increases by 17.39\%.

These results confirm that increasing the pruning rate of filters in convolutional layers has a greater impact on FLOPs reduction, while increasing the pruning rate in the FC layers affects the reduction of parameters more significantly than FLOPs.
\begin{table}[pos={b}]
\begin{center}
\fontsize{7pt}{7pt}\selectfont
\caption{Test accuracy, parameters reduction, and FLOPs reduction at various pruning rates in the convolutional and FC layers for LeNet-5 and VGG-16.}
\label{tab-Conv-FC}
\begin{tabular}{@{} llllll@{}}
\toprule
\multirow{2}{*}{Model}&\multicolumn{2}{c}{Pruning rate(\%)}&\multirow{2}{*}{Acc.(\%)}&\multirow{2}{*}{Params. $\downarrow$(\%)}&\multirow{2}{*}{FLOPs $\downarrow$(\%)}\\\cline{2-3}
&Conv&FC&&&\\
\midrule
\multirow{3}{*}{LeNet-5}&40&40&99.34&40.01&48.60\\
&40&95&99.24&91.78&56.27\\
&95&40&97.48&41.73&86.13\\
\hline
\multirow{3}{*}{VGG-16}&50&50&93.95&52.72&67.99\\
&50&90&93.50&88.32&79.03\\
&90&50&84.10&55.35&85.38\\
\bottomrule
\end{tabular}
\end{center}
\end{table}

Another notable observation is that model accuracy is generally more sensitive to filter pruning in convolutional layers compared to weight pruning in FC layers. For instance, in LeNet-5, increasing the pruning rate of the FC layers from 40\% to 95\% causes only a 0.10\% decrease in accuracy, whereas increasing the pruning rate of filters in the convolutional layers from 40\% to 95\% results in a 1.86\% accuracy drop. Similarly, for VGG-16, with 50\% pruning in both the convolutional and FC layers, the accuracy is 93.71\%. Increasing the pruning rate of the filters in convolutional layers to 90\% while keeping the FC layers at 50\% causes a more than 9.5\% drop in accuracy. In contrast, maintaining a 50\% pruning rate of filters in the convolutional layers and increasing the pruning rate of the FC layers to 90\% results in only a 0.21\% decrease in accuracy. These results can be attributed to the fact that filter pruning removes an entire batch of weights simultaneously, leading to a larger accuracy drop compared to weight pruning, which removes individual weights. Additionally, in VGG-16, the number of convolutional layers (13) significantly exceeds the number of FC layers (3). This difference in the number of layers likely contributes to the greater sensitivity of model accuracy to pruning in the convolutional layers, as pruning in these layers affects a larger portion of the model.

\begin{figure}[pos={!t}]
\centering
\includegraphics[scale=0.73]{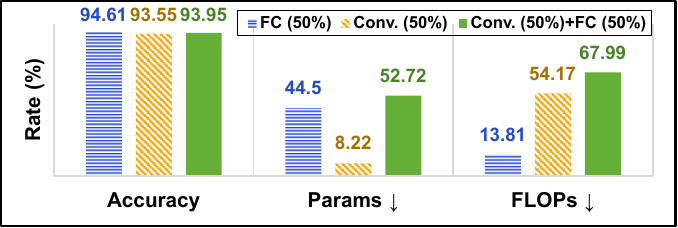}
\caption{Comparison of accuracy, parameter reduction, and FLOPs reduction under three pruning settings in VGG-16: pruning only FC layers, pruning only convolutional layers, and pruning both convolutional and FC layers. The pruning rate is fixed at 50\% for each layer type.}
\label{fig:conv-v-fc}
\end{figure}

In Fig.~\ref{fig:conv-v-fc}, we further investigate the effects of pruning convolutional and FC layers in VGG-16. Three scenarios are considered: pruning only the FC layers, pruning only the convolutional layers, and pruning both convolutional and FC layers. In all experiments, the pruning rate is fixed at 50\%, while the remaining settings are kept unchanged. 

As shown in Fig.~\ref{fig:conv-v-fc}, pruning only the FC layers achieves the highest accuracy of 94.61\%. A notable observation is that pruning both convolutional and FC layers results in an accuracy of 93.95\%, which is higher than the 93.55\% accuracy obtained by pruning only the convolutional layers. This result suggests that removing redundant parameters from the FC layers can improve model accuracy, further demonstrating the effectiveness of pruning in building highly accurate compressed DNNs. Moreover, comparing the model obtained by pruning 50\% of the weights in the FC layers with the uncompressed model further supports this observation. Specifically, pruning only the FC layers achieves an accuracy of 94.61\%, which is slightly higher than the 94.50\% accuracy of the uncompressed model reported in Table~\ref{tab-vgg-16}. 

The results also demonstrate that pruning both convolutional and FC layers achieves a higher compression ratio than the other two scenarios. Since most of the model parameters are located in the FC layers, pruning only the FC layers removes a considerably larger percentage of parameters than pruning only the convolutional layers, 44.5\% versus 8.22\%. In contrast, pruning only the convolutional layers provides a greater reduction in FLOPs, achieving a 54.17\% reduction compared with 13.81\% for pruning only the FC layers, because most FLOPs are concentrated in the convolutional layers.

\section{Discussion}\label{sec_discussion}
Because filters in a convolutional layer share weights across spatial locations and operate on all input feature maps, removing highly similar filters can have a limited impact on feature representation, as the remaining filters can preserve similar features. However, existing similarity-based pruning criteria often suffer from inherent limitations, which can negatively affect pruning effectiveness and model performance. One commonly used similarity measure is Cosine similarity \citep{ZHANG2023116, zu2023consecutive, chang2023iterative}. Cosine similarity presents a key limitation when applied to filter pruning in CNNs. Two filters may exhibit similar direction in a high Cosine similarity score, yet differ substantially in specific element-wise positions, which can significantly affect their functionality in feature extraction. Another similarity criterion, the GM criterion, has also been employed in previous studies \citep{he2019filter}. The GM criterion emphasizes overall proximity rather than element-wise similarity, potentially missing specific filter patterns. GM identifies a central point in the filter space by minimizing the sum of Euclidean distances to all other filters. Although this approach captures a notion of global centrality, it may fail to preserve critical structural characteristics. Specifically, it tends to prune central filters, those near the middle of the filter distribution, while potentially ignoring clusters of locally similar filters that reside in boundary regions farther from the median. As a result, GM-based pruning may retain redundant filters and inadequately reflect the local diversity needed for optimal feature representation.

To address these limitations, we introduce the SP criterion. Unlike Cosine similarity, which measures directional alignment between filters, SP considers the relative positioning of filters within a layer. In contrast to GM, which emphasizes global centrality based on Euclidean distance, SP represents local similarity and evaluates structural relationships across the entire filter set. Specifically, SP computes the element-wise similarity between a target filter and all other filters in the same layer, enabling the identification and preservation of filters that exhibit unique or locally significant features. This ensures that filters with distinctive characteristics are retained, thereby preserving representational diversity and enhancing model performance. Comparison of SP, GM, Cosine, and $\mathrm{L_2-norm}$ criteria in Fig.~\ref{fig:simcriterion} confirms the effectiveness of the SP criterion for filter pruning and achieving high accuracy. Moreover, the computation of the SP criterion is efficient, with a time complexity of $O(n^2\log{n})$, where $n$ denotes the number of filters in a layer.

SP is a similarity-based criterion and effectively behaves for convolutional layers, however for the FC layers, where structured pruning is typically less effective in achieving high compression efficiency, SP is not suitable. 

SP is a similarity-based criterion that is effective for convolutional layers. However, it is less suitable for FC layers, where structured pruning is generally less effective at achieving high compression rates. To address this limitation, we adopt a fine-grained magnitude-based pruning for FC layers, enabling the proposed framework to better exploit the different characteristics of convolutional and FC layers.

In many existing studies, retraining is typically carried out using fine-tuning, where the final weight values and learning rate from the original training phase are used for initialization \citep{renda2020comparing}. Reference \citep{renda2020comparing} proposes an alternative approach, known as learning rate rewinding, in which retraining begins with a learning rate taken from an earlier epoch in the training schedule, rather than the final one. In this paper, we propose an alternative strategy, learning rate refinement. Based on our experiments, we find that using a learning rate close to the final value of the original training phase yields superior results during retraining. This is likely because the final values of the unpruned weights are utilized as the initial values in the retraining phase, meaning that only minor updates are needed to guide the model toward a new optimal state. The effectiveness of our approach is demonstrated in Fig.~\ref{fig:params}, where learning rate refinement consistently outperforms both fine-tuning and learning rate rewinding.

\section{Conclusion and future Work}\label{sec_conclution}
In this work, we proposed SPRUNN, an algorithmic pruning approach that introduces the novel Structural Proximity (SP) criterion to identify candidate filters for pruning in convolutional layers. The proposed coarse-grained pruning for convolutional layers is effectively integrated with fine-grained pruning for FC layers to achieve comprehensive model compression. Moreover, SPRUNN suggested an innovative strategy, learning rate refinement, which adjusts the initial learning rate during the retraining phase to enhance model accuracy. The experimental results confirm the effectiveness of the proposed method, as SPRUNN achieves substantial compression while maintaining high accuracy across diverse DCNN architectures and datasets. Furthermore, the results demonstrate the contribution of each component of SPRUNN individually. For future work, we recommend determining the pruning rate based on the type and position of layers in a DCNN. Furthermore, investigating the influence of additional hyperparameters on the accuracy of pruned models during retraining, alongside the learning rate, would be valuable. While the SP criterion was used to assess filter similarity, it could also be extended to other structures within DNNs, such as kernels. In addition, the proposed SPRUNN framework could be implemented on FPGA platforms and evaluated using hardware-oriented metrics, such as latency, throughput, and energy consumption, to further investigate its practical deployment and acceleration capabilities.

\bibliographystyle{cas-model2-names}

\balance{}
\bibliography{cas-refs}

\end{document}